\documentclass[sigconf]{acmart}

\usepackage{times}
\usepackage{latexsym}
\usepackage{graphicx}
\usepackage{amsmath}
\usepackage{multirow}
\usepackage{adjustbox}
\usepackage{subcaption,siunitx,booktabs}
\usepackage{enumitem}

\usepackage[linesnumbered,ruled,lined]{algorithm2e}
\usepackage[switch]{lineno}
\usepackage{booktabs}
\usepackage{subcaption}
\usepackage{times}
\usepackage{epsfig}
\usepackage{pifont}

\DeclareMathOperator*{\argmax}{arg\,max}

\usepackage[T1]{fontenc}

\usepackage[utf8]{inputenc}

\acmConference[KDD '24]{SIGKDD}{August 25--29, 2024}{Barcelona, Spain}

\usepackage{microtype}
\begin{document}

\title{MAML-en-LLM: Model Agnostic Meta-Training of LLMs for Improved In-Context Learning}

\author{Sanchit Sinha}
\authornote{Work done while interning at Amazon Alexa.}
\authornotemark[0]
\email{sanchit@virginia.edu}
\affiliation{%
  \institution{University of Virginia}
  \city{Charlottesville}
  \state{VA}
  \country{USA}
}
\author{Yuguang Yue}
\email{yueyugua@amazon.com}
\affiliation{%
  \institution{Amazon Alexa}
  \city{Cambridge}
  \state{MA}
  \country{USA}
}
\author{Victor Soto}
\email{nvmartin@amazon.com}
\affiliation{%
  \institution{Amazon Alexa}
  \city{Cambridge}
  \state{MA}
  \country{USA}
}
\author{Mayank Kulkarni}
\email{maykul@amazon.com}
\affiliation{%
  \institution{Amazon Alexa}
  \city{Cambridge}
  \state{MA}
  \country{USA}
}
\author{Jianhua Lu}
\email{jianhua@amazon.com}
\affiliation{%
  \institution{Amazon Alexa}
  \city{Cambridge}
  \state{MA}
  \country{USA}
}
\author{Aidong Zhang}
\email{aidong@virginia.edu}
\affiliation{%
  \institution{University of Virginia}
  \city{Charlottesville}
  \state{VA}
  \country{USA}
}




\begin{abstract}
Adapting large language models (LLMs) to unseen tasks with in-context training samples without fine-tuning remains an important research problem. To learn a robust LLM that adapts well to unseen tasks, multiple \textit{meta-training} approaches have been proposed such as MetaICL and MetaICT, which involve meta-training pre-trained LLMs on a wide variety of diverse tasks. These meta-training approaches essentially perform in-context \textbf{multi-task fine-tuning} and evaluate on a disjointed test set of tasks. Even though they achieve impressive performance, their goal is never to compute a truly \textit{general} set of parameters. In this paper, we propose \textbf{MAML-en-LLM}, a novel method for meta-training LLMs, which can learn truly generalizable parameters that not only \textbf{performs} well on disjointed tasks but also \textbf{adapts} to unseen tasks. We see an average increase of 2\% on unseen domains in the performance while a massive 4\% improvement on adaptation performance. Furthermore, we demonstrate that MAML-en-LLM outperforms baselines in settings with \textit{limited} amount of training data on both seen and unseen domains by an average of 2\%.  Finally, we discuss the effects of type of tasks, optimizers and task complexity, an avenue barely explored in meta-training literature. Exhaustive experiments across 7 task settings along with two data settings demonstrate that models trained with MAML-en-LLM outperform SOTA meta-training approaches.  
\end{abstract}



\maketitle

\section{Introduction}
Large Language Models (LLMs) have revolutionized natural language processing (NLP) and achieved state-of-the-art performance on a variety of diverse tasks. Recently, LLMs have been demonstrated to learn \textit{in-context} \citep{wei2023larger}, without expensive and compute-intensive fine-tuning. In-context learning (ICL) involves pre-appending the target sample by carefully selected task-specific \textit{exemplars} - which act as conditioning for LLMs on that particular task. Learning in-context is an attractive proposition as its evaluation only requires inference. With no gradient updates required, ICL can potentially improve LLM generalization to new and diverse tasks with only a few examples.

 Even though out-of-the-box pre-trained LLMs show good ICL performance, multiple avenues of research have demonstrated improved ICL performance by \textit{warming-up} out-of-the-box LLMs \citep{chen2021meta,min2022metaicl}. A few model warmup techniques for improved ICL performance have been proposed recently like MetaICL \cite{min2022metaicl} and MetaICT \cite{chen2021meta}. The evaluation of such models on unseen tasks proves their efficacy in generalization. Meta-training usually involves adapting (i.e., fine-tuning) pre-trained LLMs using a diverse set of tasks, formatted as an ICL instance by pre-appending exemplars in the prompts during training. Once the model is meta-trained, the evaluation is usually performed on a distinct and disjoint set of tasks, never seen during training. The process of warming up has been dubbed as \textit{meta-training} in these approaches which borrows from research in classical machine learning literature of \textit{meta-learning} which attempts to warm up models for faster adaptation to unseen tasks.

\begin{figure}[h]
    \centering
    \includegraphics[width=0.4\textwidth]{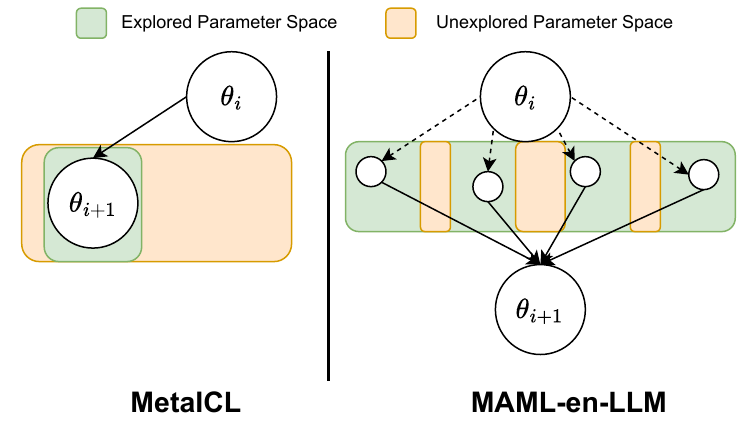}
    \caption{Visual comparison between MetaICL and \textit{MAML-en-LLM}. The figure demonstrates a single model parameter ($\theta$) update step from parameters at step $i$ - $\theta_i$ to step $i+1$ - $\theta_{i+1}$.
    The dotted lines represent the adaptation phase and the solid lines represent the update. As MetaICL does not have an explicit adaptation phase, the update happens directly and with only a limited parameter space explored. The parameter updates for a single step is calculated using only a single task. On the other hand, \textit{MAML-en-LLM} first explores a wide parameter space using multiple adapted parameters and subsequently performs the final meta-update with the second-order gradients calculated from the intermediate adapted parameters.}
    \label{fig:maml-motivation}
\end{figure} 

One of the most commonly utilized techniques for meta-learning in classical literature is Model-Agnostic Meta-Learning (MAML) proposed by \citet{finn2017model}. MAML is formulated as a two-step optimization problem where the inner loop \textit{adapts} copies of the model parameters to various diverse tasks, while the outer loop updates the initial model parameters involving a second-order gradient update calculated on the adapted parameters. Even though LLM meta-training approaches \cite{min2022metaicl,chen2021meta} are eponymous to \textit{meta-learning}, they do not utilize the two-step optimization framework for meta-training LLMs instead only adapting the model parameters continuously using a mixture of diverse tasks. For all practical purposes, the proposed meta-training approaches are analogous to \textit{multi-task fine-tuning} with added exemplars in the training sample prompts. Training models in this way does not fully exploit the parameter space and nor does it benefit from second-order gradients that guide the direction of meta-updates. In this paper, we propose \textbf{MAML-en-LLM} (pronounced \textit{Mammaliam}), a novel method for meta-training LLMs. Figure~\ref{fig:maml-motivation} visualizes the differences between our approach and the most popular state-of-the-art meta-training approach (at the time of writing) - MetaICL.  


Our main contributions are detailed below:
\begin{itemize}
    \item This paper is the first work to effectively utilize principles outlined in meta-learning literature and propose \textit{MAML-en-LLM} - a novel methodology to  meta-train LLMs, specifically for improving ICL performance.
    \item We demonstrate that \textit{MAML-en-LLM} outperforms existing meta-training approaches on ICL generalization performance on two commonly encountered settings - when training data is either limited (low resource) or abundant (high resource).
    \item We report that models trained using \textit{MAML-en-LLM} demonstrate superior performance on the challenging setting of \textit{very few-shot adaptation} to unseen domains.
    \item We present results on design considerations previously unexplored in meta-learning literature on LLMs, namely - the effect of number of exploration states (tasks), the role of optimizers, and the effect of task types on generalization. 
\end{itemize}

\section{Related Work}
\subsection{In-context Learning (ICL) in LLMs}
In-context Learning (ICL) was first proposed by \citet{brown2020language} as an extremely inexpensive alternative to regular fine-tuning. ICL requires no parameter updates as the input prompt is pre-appended with task-specific \textit{exemplars} which are examples of the specific task to be performed. LLMs have been shown to perform exceptionally well when prompted with a few examples of the task pre-appended in the prompt \cite{wei2023larger}. This behavior was first studied for LLMs in GPT-3 \cite{radford2019language}. Subsequent studies have shown that they can even solve complex problems like math \cite{radford2019language} and reasoning \cite{wei2022chain}. Why LLMs are so adept at in-context learning remains an open research topic that is attracting attention; for example \citet{akyurek2022learning} compares their behavior to linear models. Empirical approaches like \cite{yoo2022ground} and \cite{lu2021fantastically} have demonstrated that types of exemplars are important for ICL performance. 

\subsection{Meta-Learning for Generalization}
One of the most popular approaches for meta-learning was first proposed by \cite{finn2017model}
as Model Agnostic Meta-Learning (MAML). The goal of MAML was to \textit{warmup} pre-trained models over a diverse set of tasks using an inner-outer minimization algorithm to learn a general set of parameters that can be adapted to new tasks using only a few examples (few-shot).

The inner loop usually focuses on a set of tasks, while the outer loop utilizes the learned parameters and the second-order gradient information from the inner loop to update the model, in turn capturing the direction of gradient updates. Usually, MAML requires the outer loop to perform a second-order gradient update which results in a higher computational burden. Even though MAML is effective, it is not without serious issues; for example, it is known to suffer from memorization \cite{yin2019meta}, which entails the model repeating the data shown to it during training without learning anything. MAML has also been shown to be sensitive to training hyper-parameters and complex strategies have been proposed to enforce stability during its training process \cite{antoniou2018train}. MAML has also been used with some success for training language models \cite{wang2021variance, deb2022boosting, lux2022language, kirsch2022general} on particular tasks.

\subsection{Meta-training for improving ICL}
To improve the ICL performance of out-of-box pre-trained models, several meta-training approaches have been proposed. We discuss the two seminal works - MetaICT \cite{chen2021meta} and MetaICL \cite{min2022metaicl} here. MetaICT uses BinaryCLFs and LAMA datasets to create tasks while also pre-appending human-generated instructions to each task. MetaICL on the other hand uses a wide variety of disjoint tasks with no human-generated instructions to meta-train pre-trained LLMs. Both approaches update the model parameters on a batch from training tasks continuously - making them similar to fine-tuning. A recent related work \cite{qin2023learning} attempts to utilize MAML to train LLMs for improving Prompt Tuning. However, the work is significantly different from ours - as generalization is performed on learned soft embedding of tokens and not on model parameters - a vastly different objective from ours which is improving ICL performance. 

\noindent \textbf{Comparision to existing works.} We discuss the salient differences with MetaICL first. MetaICL discards principles of meta-learning and effectively only uses multi-task fine-tuning to meta-train their models. Next, MetaICT follows an identical training process to MetaICL. Even though MetaICT attempts to compare MetaICT to MAML, they only utilize the first-order approximation and a single task to update the model parameters in the inner optimization step possibly due to not conducting proper investigation into a myriad of sources of errors like optimizers and exploration states. Our work is significantly different wherein, we use second-order gradients to meta-train models and also provide exhaustive results and discussions around utilizing MAML-en-LLM for various types of tasks. 

\begin{figure*}
    \centering
    \includegraphics[width=0.85\textwidth]{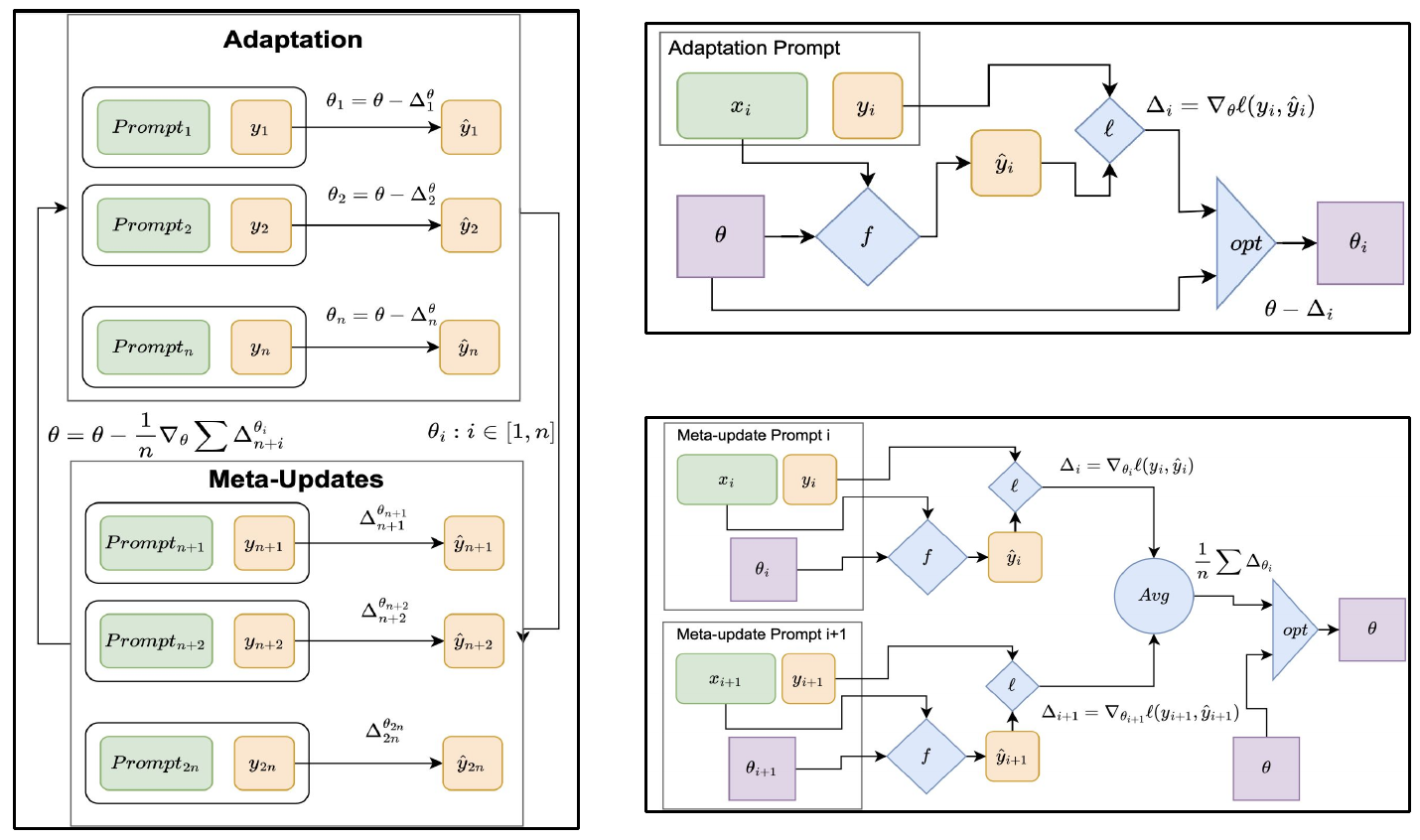}
    \caption{Schematic figure demonstrating MAML training for meta-training LLMs. MAML is a bi-level optimization framework with an inner update (adaptation) and an outer update (meta-update). In the figure, the green cells represent the input samples (prompts), the yellow cells represent task labels, blue components represent functions and purple boxes represent model parameters. Multiple task batches are utilized to compute a set of adapted parameters (equal to the number of tasks) represented by $\theta_{i}$. The outer update utilizes the adapted parameters to compute second-order gradients ($\Delta_{i}^{\theta_i}$) using a separate set of task batches. The final meta-update updates the unadapted parameter $\theta$ with an average of second-order gradients ($\Delta_{i}^{\theta_i}$) to compute the next set of updated parameters.}
    \label{fig:schematic-maml}
\end{figure*}
\section{Methodology}
\label{sec:method}
In this section, we first go over the standard meta-training procedure for improving ICL performance of LLMs by discussing the problem setup. Subsequently, we provide a mathematical overview of current SOTA approaches and our proposed approach. Next, we provide finer details for \textit{MAML-en-LLM} and its various components - task adaptation and aggregated meta-update phases in addition to detailed optimization perspective. 

\subsection{Meta-training: Problem Statement}
\label{sec:problem-setup}
In-context learning is an inference-only method, where exemplars from the same tasks with expected output values are provided in the prompt as conditioning for the LLM. During meta-training, a similar setup is followed where for a given train sample $(x_i,y_i)$, belonging to a task $\mathbb{C}$, $k$ samples $(x_1,y_1)$,$(x_2,y_2)$,...,$(x_k,y_k)$ from the same task $\mathbb{C}$ are sampled. The $k$ samples are pre-appended in the prompt with the final train sample $x_i$. The target label $y_i$ is used to calculate the loss and the pre-trained LLM ($f$) is meta-trained over all available training tasks $\mathbb{C}$ with standard classification training objectives ($\ell$). Mathematically, a parameter update step can be represented as:
\begin{align}
    \theta = \theta - \nabla_{\theta}~\ell(f(x_1,y_1,x_2,y_2,...x_k,y_k,x_i),y_i) ~~~~\forall(x_i,y_i)\in\mathbb{C}
\end{align}
We provide more detail about the exact update methodology in the next section.

\subsection{Model Agnostic Meta-Learning for LLMs (\textit{MAML-en-LLM})}
Note that the present state-of-the-art meta-training methods like MetaICL\cite{min2022metaicl} and MetaICT\cite{chen2021meta} perform optimization in line with multi-task finetuning. Mathematically, MetaICL and MetaICT solve a single optimization objective:
\begin{equation}
    min~\theta ~~\sum_{\mathcal{T}_i \sim p(\mathcal{T})}\mathcal{L}_{\mathcal{T}_i}(f_{\theta})
\end{equation}
On the other hand, \textit{MAML-en-LLM} aims to learn a generalizable set of parameters by training a model on a diverse variety of tasks with two distinct phases - an inner phase which controls the extent of exploration - adaptation and an outer phase which controls the magnitude of the update - the meta-update. As a consequence, \textit{MAML-en-LLM} can be represented as a bi-level optimization with an inner update step and an outer (meta) update step where it solves a dual optimization training objective given by: 
\begin{equation}
    min~\theta ~~\sum_{\mathcal{T}_i \sim p(\mathcal{T})}\mathcal{L}_{\mathcal{T}_i}(f_{\theta - \nabla_\theta \mathcal{L}_{\mathcal{T}_i} (f_{\theta})})
\end{equation}

\subsubsection{\large \textbf{Task Adaptation}}
The inner optimization adapts the parameters to a set of tasks. To achieve this $n$ tasks are sampled where each task is a set of size $k$. The model is adapted on each task by performing gradient descent on the original parameters for $k$ steps also known traditionaly as \textit{support set} $S$. The number of elements $k$ in the \textit{support set} is equal to the number of steps for adaptation using gradient descent. Intuitively, the higher the number of tasks $n$, the greater the exploration of the parameter space. The adapted parameters are calculated using Equation~\ref{eq:maml-inner}. Mathematically, the inner loop adapts to a distribution of tasks $\mathcal{T}_i \sim p(\mathcal{T})$ where $p(\mathcal{T})$ represents a probability distribution over diverse tasks. Note that a task represents a randomly sampled batch of training prompts as discussed in \citet{min2022metaicl}.
\begin{equation}\label{eq:maml-inner}
    \theta_i \leftarrow \theta - \alpha\nabla_{\theta}\ell_{\mathcal{T}_i}(f_{\theta})
\end{equation}
where $\mathcal{T}_i \sim p(\mathcal{T})$ is sampled from all tasks and $\theta$ represents the model parameters,  $\alpha$ is the learning rate of the adaptation step and $\ell$ is the Cross Entropy Loss.

\subsubsection{\large \textbf{Aggregated Meta-updates}}
The outer update utilizes a distinct \textit{query set} $Q (\perp S)$ to calculate the gradients with respect to the adapted parameters. The size of the query set is kept the same as the support set (i.e. $k$ = $\|Q\|$ = $\|S\|$). In classic MAML literature, the same query set is usually used to calculate the meta-update. However \textit{MAML-en-LLM} utilizes $n$ different tasks to perform the meta-updates. For each task the calculated gradients (using the query set) are collected and averaged to perform a second-order gradient update on the original unadapted parameter as per Equation~\eqref{eq:maml-outer}. The outer optimization performs the meta-update on the unadapted parameter based on the second-order gradient updates of the adapted set of parameters $\theta_i's$. Mathematically the second-order update can be represented as:

\begin{equation}\label{eq:maml-outer}
    \theta \leftarrow \theta - \beta\nabla_{\theta}\Sigma_{\mathcal{T}_i \sim p(\mathcal{T})}\ell_{\mathcal{T}_i}(f_{\theta_i}) 
\end{equation}

\noindent where $\beta$ is the learning rate for the outer optimization step.

\subsection{Shared Adaptive Optimizer Moments}
\label{sec:method-optim}
One of the most significant differences between \textit{MAML-en-LLM} and MetaICL is the presence of a dual optimization problem in \textit{MAML-en-LLM}. The dual optimization problem poses unique challenges in LLMs where the choice of optimizers affects generalization drastically. Note that typically, adaptive optimizers (like AdamW) are preferred over stateless optimizers (like SGD with momentum). For a typical adaptive optimizer, given gradients $g_{t}$ at step $t$, observations sampled from a batch $B$, hyperparameters $\beta_1$ and $\beta_2$, two moving averages are calculated - the first moment $m_t$ and second moment $v_t$ as follows:
\begin{align*}
    m^B_t \leftarrow (\beta_1*m_{t-1} + (1-\beta_1)*g_t)/(1-\beta_1^t) \\
    v^B_t \leftarrow (\beta_2*v_{t-1} + (1-\beta_2)*g_t^2)/(1-\beta_2^t)
\end{align*}

Note that the optimizer in the inner update is re-initialized after every meta-update, implying that the moments are re-initialized as well. This significantly changes the landscape of optimization, making inner gradient updates contribute disproportionately more as the training continues for a longer duration, especially near the minima. To alleviate this problem, we propose optimizer \textit{parameter sharing} between the inner and outer optimizers - i.e., constantly updating shared moving averages between optimizers where only a single set of optimizer parameters are updated.


\begin{algorithm}[h]
    \SetAlgoLined
    \DontPrintSemicolon
    \caption{MAML-en-LLM}\label{algo:maml-en-llm}
        \SetKwInOut{Input}{Input}
        \Input{Training samples:$X$, Task Labels:$Y$, Training Corpus:~$\{x_i\in X$;~$y_i\in Y\}$ , Steps:~$N$, Model $f$, Pre-trained Model Parameters:~$\theta_{pt}$, Learning Rates:~$\alpha,\beta$, Moment hyperparameters $\beta_1, \beta_2$, Size of support and query set $n$, Loss function $\ell$ usually the cross entropy loss }
        \SetKwInOut{Output}{Output}
        \Output{Meta-trained model parameters $\theta$}
        $ \theta \gets \theta_{pt}$\;
        $t \gets 0,m,v \gets 0$\;
        \For{$t~\in~1,2,3,...,N$}
        {
            \text{Sample Support set of size $n$, $\{x_i\}^n,\{y_i\}^n \in \{X,Y\}$}
            
            \For{$i~\in~1,2,3,..n$}
            {
                $g_i(t) = \nabla_{\theta}\ell(f_\theta(x_i),y_i)$
                
                $m \leftarrow (\beta_1*m + (1-\beta_1)*g_i(t))/(1-\beta_1^t)$

                $v \leftarrow (\beta_2*v + (1-\beta_2)*(g_i(t))^2)/(1-\beta_2^t)$

                $\theta_i \gets \theta - (\alpha * m / \sqrt{v} + |\theta|_2) $
                
            }
            \text{Sample Query set of size $n$, $\{x_i\}^n,\{y_i\}^n \in \{X,Y\}$}
            
            \For{$i~\in~1,2,3,..n$}
            {
                $g'(t) = \frac{1}{n}~*~\sum_k~\ell(f_{\theta_i}(x_i),y_i)$

                $m \leftarrow (\beta_1*m + (1-\beta_1)*g'(t))/(1-\beta_1^t)$

                $v \leftarrow (\beta_2*v + (1-\beta_2)*(g'(t))^2)/(1-\beta_2^t)$

                $\theta \gets \theta - (\beta * m / \sqrt{v} + \frac{1}{k}*\sum_k |\theta_i|_2) $
                  
            }      
        }
\end{algorithm}


\subsection{Consolidated Meta-Training using MAML-en-LLM}
Figure~\ref{fig:schematic-maml} provides a schematic overview of the consolidated training approach and Algorithm~\ref{algo:maml-en-llm} details the exact MAML-en-LLM training procedure. As we utilize $n$ tasks for calculating the adapted parameters and subsequently utilize $k$ tasks for adaptation steps and calculating meta-updates, we represent our \textit{MAML-en-LLM} terminology by \textit{MAML-2k-n}. Hence, if the size of the support and query set is 1 and the number of tasks during adaptations are also 1, the \textit{MAML-en-LLM} setting will be represented as \textbf{MAML-2-1}. It is intuitive to see that the meta-update frequency can easily be calculated as $2kn$, hence for MAML-2-1, the frequency of meta-updates is $2$. Similarly, for \textbf{MAML-2-4}, the frequency of meta-updates is $8$.
\section{Experiments}
\subsection{Dataset Description}
We utilize two datasets with a wide diversity of tasks - \textsc{CrossFit} \cite{ye2021crossfit} and \textsc{UnifiedQA} \cite{khashabi2020unifiedqa} consisting of total of 142 distinct tasks. We use the exact same task splits used by MetaICL \cite{min2022metaicl}, however due to active developments, we utilize latest versions of tasks. All the tasks fall in the following 4 categories with increasing complexity: text classification, natural language inference, question answering, and paraphrasing. The training tasks and the testing tasks are ensured to be disjointed. The testing tasks consist of two types of tasks - tasks with similar domains in training set and sampled from unseen domains in training set. The statistics of the dataset splits are detailed in Table~\ref{tab:dataset-split}.

\begin{table}[t]
\centering
\resizebox{0.48\textwidth}{!}{
\begin{tabular}{c|c|c|cc}
\hline
   \textbf{Train Setting}                & \textbf{Train} &      \textbf{Test Setting}  & \textbf{Test} & \textbf{Unseen} \\
                   \hline
HR                 &    61   & LR                              &   26   &     4   \\
\hline
Classification     &   43    & \multirow{2}{*}{Classification} &   \multirow{2}{*}{20}   &     \multirow{2}{*}{4}   \\
Non-Classification &   37    &                                 &      &        \\
\hline
QA                 &     37  & \multirow{2}{*}{QA}             &   \multirow{2}{*}{22}   &    \multirow{2}{*}{4}    \\
Non-QA             &    33   &                                 &      &        \\
\hline
Non-NLI            &   55    &   NLI                              &  8    &   1   \\
\hline
Non-Paraphrase     &   59    & Paraphrase                      &   4   &  1     \\
\hline
\end{tabular}
}
\caption{Statistics of number of train/test tasks in the 7 meta-training settings considered for evaluation. The number of tasks in the Unseen domain is listed in the last column. Dataset split is identical to \citet{min2022metaicl}}
\label{tab:dataset-split}
\end{table}

\subsection{Training Details}
\label{sec:training-deets}
We utilize a pre-trained GPT-2 Medium \cite{radford2019language} consisting of 355 million parameters for all our experiments. We consider both the standard models and Noisy channel models \cite{min2021noisy} for evaluation of all experiments. Training is performed on 8 Tesla V100 GPUs using Pytorch and pre-trained model checkpoints are taken from HuggingFace. Training time for MetaICL models for 50k steps is about 5 hours, while for \textit{MAML-en-LLM} models is about 12 hours.

\begin{figure}[h]
    \centering
    \includegraphics[width=0.35\textwidth]{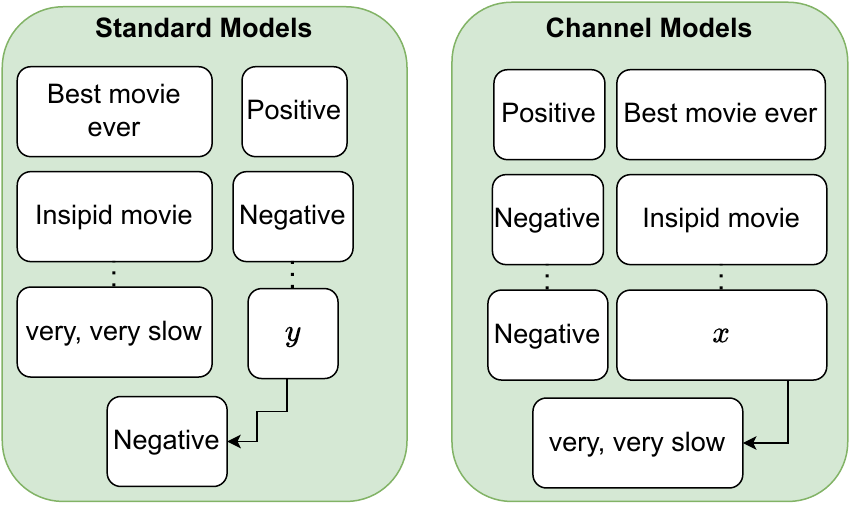}
    \centering
    \includegraphics[width=0.35\textwidth]{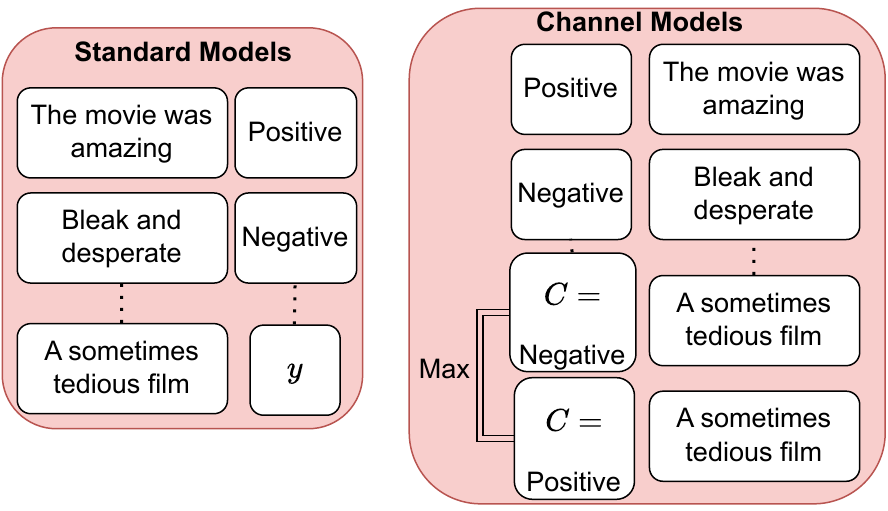}
    \caption{Example training and test prompts for Standard (Left) and Channel (Right) Models. The training procedure of Channel models learns to predict the sample, conditioned on its true label. During inference, channel models predict the target sample itself conditioned on all possible labels for the task. In this example, the task is Sentiment Analysis and hence has only two labels Positive and Negative.}
    \label{fig:prompt-examples}
\end{figure}

\noindent \textbf{Example prompt structure}.
We first visualize the prompt structure during training of both standard and channel models (Figure~\ref{fig:prompt-examples} (Green). Next, we visualize the prompt structure during ICL inference of both standard and channel models (Red). The $Max$ operator selects the label $C$ with the max probability,

\subsection{Hyper-parameter settings}
\noindent \textbf{Training.} For both MetaICL and MAML models, the training sequence length is fixed at 1024. The batch size is fixed at 1. For MetaICL baselines, we utilize the learning rate of $1$x$10^{-5}$. As MAML consists of an inner ($\alpha$) and outer learning rate ($\beta$), we utilize the same learning rate of $1$x$10^{-5}$ for both $\alpha$ and $\beta$. For MAML, the size of support and query sets is chosen as $1$ (number of batches). The number of tasks is chosen as $1$ (MAML-2-1) and $4$ (MAML-2-4). This implies the frequency of meta-updates is $2$ and $8$ respectively. We train both paradigms of models for 50k steps. For both approaches, AdamW is used as an optimizer. As MAML models require an inner and outer optimizer, we utilize AdamW for both optimizers with identical hyper-parameters. As every AdamW's optimization step depends on the last gradient update, we copy the inner optimizer's last gradient state to the outer optimizer for every meta-update. The training seed is set as 100. More details on optimizers can be found in the Appendix\ref{optimizer_appendix}.

\noindent \textbf{Inference.} We utilize identical evaluation framework as \cite{min2022metaicl} for fair comparisons. The test sequence length is fixed at 256 or 16 exemplars whichever is lower. The batch size is fixed at 16 samples during inference. 

\begin{table*}[t]
    \centering
   \begin{subtable}[h]{0.95\textwidth}
    \centering
    \begin{adjustbox}{max width=0.95\textwidth}
    \begin{tabular*}{1.15\textwidth}{cccccccc}
    \hline
     Method        & \multicolumn{1}{p{2cm}}{\centering HR $\rightarrow$ LR} & \multicolumn{1}{p{2cm}}{\centering Class \\ $\rightarrow$ Class} & \multicolumn{1}{p{2cm}}{\centering Non-Class \\ $\rightarrow$ Class} & \multicolumn{1}{p{2cm}}{\centering QA \\ $\rightarrow$ QA} & \multicolumn{1}{p{2cm}}{\centering Non-QA \\ $\rightarrow$ QA} & \multicolumn{1}{p{2cm}}{\centering Non-NLI \\ $\rightarrow$ NLI} & \multicolumn{1}{p{2cm}}{\centering Non-Para \\ $\rightarrow$ Para}  \\
    \hline
    No ICL   & 27.71 & 27.71 & 27.71 & 44.06 & 44.06 & 33.5 & 35.23\\
    Raw LM   & 29.12/25.81 & 29.12/25.81 & 29.12/25.81  & 45.43/44.68 & 45.43/44.68 & 42.59/34.01 & 41.43/34.72\\
    \hline
    MetaICL  & 42.54/38.97 & 38.14/36.25 & \textbf{36.08}/32.56 & 59.87/58.43 & 38.30/35.93 & 78.85/73.2 & \textbf{59.86}/54.92 \\
    \hline
    MAML-2-1 & 34.59/33.57 & \textbf{42.94}/41.31 & 31.55/26.41 & \textbf{61.99}/61.25 & 34.93/32.81 & \textbf{80.61}/77.11 & 33.32/32.22\\
    MAML-2-4 & \textbf{43.16}/42.11 & 40.65/39.99 & 32.47/27.61 & 59.37/58.75 & \textbf{40.49}/39.06 & 60.47/52.8 & 34.05/34.05\\
    \hline
    \hline
    Channel No ICL   & 30.68 & 30.68 & 30.68 & 45.93 & 45.93 & 33.29 & 41.09\\
    Channel Raw LM   & 42.66/39.81 & 42.66/39.81 & 42.66/39.81 & 44.24/41.87 & 44.24/41.87 & 38.50/34.7 & 53.03/48.41 \\
    \hline
    Channel MetaICL  & 48.71/47.27 & \textbf{49.46}/47.49 & 44.08/43.35 & \textbf{57.37}/55.93 & \textbf{48.49}/46.87 & 55.71/45.07 & 48.49/46.87 \\
    \hline
    Channel MAML-2-1 & \textbf{51.19}/48.54 & 48.65/47.01 & 46.45/44.52 & 55.93/54.06 & 46.81/45 & 64.55/60.18 & \textbf{59.15}/56.02\\
    Channel MAML-2-4 & 50.93/47.52 & 49.04/47.79 & \textbf{46.83}/44.86 & 55.62/54.06 & 47.06/45 & \textbf{64.99}/59.99 & 57.97/54.17\\
    \hline
    \end{tabular*}
    \end{adjustbox}
    \caption{Performance on unseen tasks utilizing \textit{complete data setting.} }
    \label{tab:performance-unseen-data}
    \end{subtable}
    \hfill
    \begin{subtable}[h]{0.95\textwidth}
    \centering
    \begin{adjustbox}{max width=0.95\textwidth}
    \begin{tabular*}{1.15\textwidth}{cccccccc}
    \hline
     Method        & \multicolumn{1}{p{2cm}}{\centering HR $\rightarrow$ LR} & \multicolumn{1}{p{2cm}}{\centering Class \\ $\rightarrow$ Class} & \multicolumn{1}{p{2cm}}{\centering Non-Class \\ $\rightarrow$ Class} & \multicolumn{1}{p{2cm}}{\centering QA \\ $\rightarrow$ QA} & \multicolumn{1}{p{2cm}}{\centering Non-QA \\ $\rightarrow$ QA} & \multicolumn{1}{p{2cm}}{\centering Non-NLI \\ $\rightarrow$ NLI} & \multicolumn{1}{p{2cm}}{\centering Non-Para \\ $\rightarrow$ Para}  \\
    \hline
    No ICL   & 27.71 & 27.71 & 27.71 & 44.06 & 44.06 & 33.5 & 35.23\\
    Raw LM   & 29.12/25.81 & 29.12/25.81 & 29.12/25.81  & 45.43/44.68 & 45.43/44.68 & 42.59/34.01 & 41.43/34.72\\
    \hline
    MetaICL  & \textbf{43.78}/41.21 & 37.22/34.06 & \textbf{35.31}/31.64 & 51.24/49.68 & \textbf{44.62}/41.87 & 33.16/33.16 & 38.66/32.59 \\
    \hline
    MAML-2-1 & 41.30/39.08 & \textbf{38.67/36.96} & 34.02/31.78 & 52.93/51.24 & 43.68/41.87 & 53.57/42.07 & 34.02/33.98\\
    MAML-2-4 & 42.29/40.18 & 38.34/36.95 & 34.04/31.94 & \textbf{53.49}/52.81 & 41.93/38.43 & \textbf{57.52}/55.09 & \textbf{39.21}/38.15\\
    \hline
    \hline
    Channel No ICL   & 30.68 & 30.68 & 30.68 & 45.93 & 45.93 & 33.29 & 41.09\\
    Channel Raw LM   & 42.66/39.81 & 42.66/39.81 & 42.66/39.81 & 44.24/41.87 & 44.24/41.87 & 38.50/34.7 & 53.03/48.41 \\
    \hline
    Channel MetaICL  & 45.06/42.97 & \textbf{46.98}/44.52 & 43.69/42.31 & 55.36/53.12 & \textbf{50.62}/48.75 & 53.57/34.39 & \textbf{53.72}/49.73   \\
    \hline
    Channel MAML-2-1 & \textbf{48.75}/46.44 & 45.9/44.08 & \textbf{45.4}/43.84 & 55.62/54.37 & 49.62/48.12 & \textbf{54.16}/37.22 & 52.74/48.06\\
    Channel MAML-2-4 & 46.01/43.53 & 45.76/44.08 & 45.27/44.36 & \textbf{55.80}/ 54.06 & 49.55/48.43 & 54.02/37.1 & 52.76/47.82\\
    \hline
    \end{tabular*}
    \end{adjustbox}
    \caption{Performance on unseen tasks utilizing \textit{limited data setting.}}
    \label{tab:performance-unseen-small-data}
    \end{subtable}
    
    \caption{For both tables, Rows 1-3 represent the baselines and Rows 4 and 5 represent the performance of MAML settings. Similarly, Rows 6-8 represent baselines using the Channel training/inference and Rows 9 and 10 represent MAML settings on Channel models. Numbers are represented as X/Y where X represents the average performance and Y represents the worst-case performance. The entries in \textbf{bold} are the best-performing models.}
    \label{tab:performance-unseen-tasks}
\end{table*} 

\subsection{Comparison of Baselines and Replication}
We compare our proposed methodology against a variety of different models and prompt settings. We utilize 2 types of models - standard models and noisy channel models as proposed by \citet{min2021noisy}. For standard models, the training and inference procedure for a given input sample $x_i,y_i$ and exemplars $(x_1,y_1,...x_k,y_k)$ is as follows (Refer problem setup in Section~\ref{sec:problem-setup}):   
\begin{equation}
    \label{eq:standard-train-infer}
     \theta = \theta - \nabla_{\theta}~\ell(f(x_1,y_1,x_2,y_2,...x_k,y_k,x_i),y_i)
\end{equation}
\begin{equation}
    y_i = \argmax P(y|x_1,y_1,x_2,y_2,...x_k,y_k,x_i)
\end{equation}
where $\ell$ is the Cross Entropy loss.
On the other hand, noisy-channel models (referred to as Channel models from now on) treat the training and inference procedure as a generative problem rather than a classification task. During training, the labels and prompts are flipped and reordered, and the target label is appended in the prompt itself, while the training instance is treated as the model's generated output. Mathematically, the training procedure optimizes the following:
\begin{equation}
     \label{eq:channel-train}
     \theta = \theta - \nabla_{\theta}~\ell(f(y_1,x_1,y_2,x_2,...y_k,x_k,y_i),x_i)
\end{equation}
During inference, all possible target labels are considered and appended in the prompt, and the prompt with the highest probability is treated as the correct label. 
\begin{equation}
    \label{eq:channel-infer}
    y_i = \argmax_{c \in C}P(x_i|y_1,x_1,y_2,x_2,...y_k,x_k,c)
\end{equation}
where $C$ represents all possible labels for the task.
For both models, we consider 3 different settings based on the prompt structure and the model used. We detail the settings below. 
\begin{itemize}
    \item \textbf{No ICL}: The prompt only consists of the target instance($x_i$) and estimates target label ($y_i$).  
    \item \textbf{RawLM}: The prompt consists of exemplars ($x_1$,$y_1$,..,$x_{n}$,$y_{n}$) followed by the target sample ($x_i$) and estimates the target label ($y_i$).  The model used is a pre-trained out-of-box LLM. Note, for Channel models, the prompt structure is flipped as shown in Equation~\ref{eq:channel-infer}.
    \item \textbf{MetaICL}: We replicate the exact MetaICL setting by training the model identical to \cite{min2022metaicl}. Please refer to Section~\ref{sec:training-deets}, for more information.
\end{itemize}
\noindent \textbf{Replication of MetaICL.} We replicate the training procedure of MetaICL and Channel MetaICL models on the latest versions of the datasets and on the GPT-2 Medium models. Differently from the provided approach, we do not utilize 8-bit optimization and mixed precision training. Our replicated results agree with the reported results on all the data splits in the paper and even perform better than the reported results on GPT-2 Medium (Table 11 in \cite{min2022metaicl}). 

\begin{table*}[t]
\begin{subtable}[h]{0.95\textwidth}
\centering
\begin{adjustbox}{max width=0.95\textwidth}
\begin{tabular*}{1.15\textwidth}{cccccccc}
\hline
Method        & \multicolumn{1}{p{2cm}}{\centering HR $\rightarrow$ LR} & \multicolumn{1}{p{2cm}}{\centering Class \\ $\rightarrow$ Class} & \multicolumn{1}{p{2cm}}{\centering Non-Class \\ $\rightarrow$ Class} & \multicolumn{1}{p{2cm}}{\centering QA \\ $\rightarrow$ QA} & \multicolumn{1}{p{2cm}}{\centering Non-QA \\ $\rightarrow$ QA} & \multicolumn{1}{p{2cm}}{\centering Non-NLI \\ $\rightarrow$ NLI} & \multicolumn{1}{p{2cm}}{\centering Non-Para \\ $\rightarrow$ Para}  \\
\hline
No ICL  & 33.09 & 33.3 & 33.3 & 38.85 & 38.85 & 25.89 & 33.44 \\
Raw LM  & 36.68/35.98 & 34.95/34.24 & 34.95/34.24 & 39/38.64 & 39/38.64 & 32.91/31.41 & 37.26/35.32\\
\hline
MetaICL & \textbf{42.42}/41.25 & 42.52/42.11 & \textbf{42.21}/40.88 & 41.48/41.25 & \textbf{36.71}/35.97 & 50.03/46.58 & 33.1/33.1 \\
\hline
MAML-2-1 & 41.88/41.43 & 41.88/41.08 & 40.09/38.69 & 42.38/42.22 & 20.36/18.98 & \textbf{50.05}/46.81  & 16.39/16.04\\
MAML-2-4 & 41.6/40.81 & \textbf{42.82}/42.01 & 40.46/39.29 & \textbf{42.56}/42.41  & 36.13/35.32 & 44.71/42.42 & \textbf{38.64}/38.14\\
\hline
\hline
Channel No ICL & 36.59 & 33.98 & 33.98 & 38.28 & 38.28 & 33.31 & 46.16   \\
Channel Raw LM & 41.95/40.83 & 45.31/45.09 & 45.31/45.09 & 39.97/39.51 & 39.97/39.51 & 38.83/37.65 & 45.2/44.38    \\
\hline
Channel MetaICL  &  46.25/45.42 & 50.35/49.71 & 49.03/48.02 & \textbf{42.75}/42.54 & \textbf{40.46}/40.08 & 48/47.1 & 51.30/49.76 \\
\hline
Channel MAML-2-1 & 48.10/47.16 & 51.08/50.46 & 50.31/49.87 & 42.67/42.43 & 40.13/39.93 & 52.38/51.22 & \textbf{54.18}/52.61 \\
Channel MAML-2-4 & \textbf{48.03}/47.26 & \textbf{51.13}/50.49 & \textbf{50.49}/49.9 & 42.51/42.21 & 40.03/39.77 & \textbf{52.43}/51.43 & 53.34/51.91\\
\hline
\end{tabular*}
\end{adjustbox}
\caption{Performance on all tasks utilizing \textit{complete data setting.} }
\label{tab:performance-all-data}
\end{subtable}
\hfill
\begin{subtable}[h]{0.95\textwidth}
\centering
\begin{adjustbox}{max width=0.95\textwidth}
\begin{tabular*}{1.15\textwidth}{cccccccc}
\hline
Method        & \multicolumn{1}{p{2cm}}{\centering HR $\rightarrow$ LR} & \multicolumn{1}{p{2cm}}{\centering Class \\ $\rightarrow$ Class} & \multicolumn{1}{p{2cm}}{\centering Non-Class \\ $\rightarrow$ Class} & \multicolumn{1}{p{2cm}}{\centering QA \\ $\rightarrow$ QA} & \multicolumn{1}{p{2cm}}{\centering Non-QA \\ $\rightarrow$ QA} & \multicolumn{1}{p{2cm}}{\centering Non-NLI \\ $\rightarrow$ NLI} & \multicolumn{1}{p{2cm}}{\centering Non-Para \\ $\rightarrow$ Para}  \\
\hline
No ICL & 33.09 & 33.3 & 33.3 & 38.85 & 38.85 & 25.89 & 33.44   \\
Raw LM   & 36.68/35.98 & 34.95/34.24 & 34.95/34.24 & 39/38.64 & 39/38.64 & 32.91/31.41 & 37.26/35.32\\
\hline
MetaICL & 39.52/38.71 & 38.26/37.59 & 41.13/40.51 & 40.04/39.33 & \textbf{38.85}/38.23 & 31.51/28.49 & 34.29/32.78 \\
\hline
MAML-2-1 & \textbf{39.86}/38.82 & 41.68/39.77 & \textbf{42.31}/41.44 & 40.6/40.21 & 38.09/37.37 & \textbf{38.61}/33.41 & 33.14/33.14\\
MAML-2-4 & 39.27/38.4 & \textbf{42.24}/40.91 & 41.92/40.67 & \textbf{40.7}/40.1 & 38.28/37.45 & 37.69/32.92 & \textbf{34.43}/34.17 \\
\hline
\hline
Channel No ICL & 36.59 & 33.98 & 33.98 & 38.28 & 38.28 & 33.31 & 46.16\\
Channel Raw LM & 41.95/40.83 & 45.31/45.09 & 45.31/45.09 & 39.97/39.51 & 39.97/39.51 & 38.83/37.65 & 45.2/44.38    \\
\hline
Channel MetaICL  & 45.09/43.83 & \textbf{48.36}/47.3 & 46.77/46.23 & 42.20/41.79 & 40.60/40.14 & 44.51/42.7 & 48.64/47.29\\
\hline
Channel MAML-2-1 & \textbf{46.16}/45.3 & 48.27/47.89 & 47.27/46.49 & \textbf{42.46}/42.19 & 40.64/40.25 & 46.51/45.32 & \textbf{50}/47.83 \\
Channel MAML-2-4 & 45.054/43.93 & 47.89/47.55 & \textbf{47.24}/46.54 & 42.32/42.11 & \textbf{40.74}/40.48 & \textbf{46.74}/45.05 & 49.606/48.33\\
\hline
\end{tabular*}
\end{adjustbox}
\caption{Performance on all tasks utilizing \textit{limited data setting.}}
\label{tab:performance-all-small-data}
\end{subtable}
\caption{For both tables, we report the performance numbers on all the tasks. Rows 1-3 represent the baselines and Rows 4 and 5 represent the performance of MAML settings with 1 and 4 tasks respectively. Similarly, Rows 6-8 represent baselines using the Channel training/inference and Rows 9 and 10 represent MAML settings on Channel models. The numbers are represented as X/Y where X represents the average performance and Y represents the worst case performance. The entries in \textbf{bold} are best-performing models.}
\label{tab:performance-all-tasks}
\end{table*}

\subsection{Evaluation Criterion and Metrics} 
To quantify the performance of classification tasks, macro-F1 score is utilized (Refer \cite{min2022metaicl} for details) which is suitable for settings with class imbalances. For all other tasks prediction accuracy is used instead. 
We report the average and worst-case performance on five random seeds (identical to \cite{min2022metaicl}). To compare performances among various methods for the same data setting, we report the \textbf{win-rates} of \textit{MAML-en-LLM} models. We consider a `win' for all cases where both the average and worst case performances of \textit{MAML-en-LLM} models outperform their counterparts and are significant (discussed later). The numbers in bold represent the best-performing methods for each data setting.

\section{Results and Discussion}
\subsection{Experiment-1: Generalization Performance}
We evaluate our methodology on two commonly encountered data settings in practice. These settings are useful to demonstrate the efficacy of our method on both high and low-resource settings.

\noindent \textbf{Complete Data Setting (High Resource).} Comprised of the entire training set from each task dataset. We utilize the exact same task and data splits as MetaICL\cite{min2022metaicl}. The statistics for training and testing data are detailed in Table~\ref{tab:dataset-split}.

\noindent \textbf{Limited Data Setting (Low Resource).} We sample 10\% of training data from each task dataset utilizing the same dataset seeds. To combat the bias introduced during sampling, we ensure the proportion of labels in the sampled data is the same as in the complete data setting, i.e., the sampling is equally stratified (Table~\ref{tab:dataset-split}).

To compare the generalization performance of \textit{MAML-en-LLM} and MetaICL, we evaluate first on test tasks in unseen training domains and next on all test tasks. As ICL is sensitive to the selected \textit{exemplars} in the prompt,  we consider five seeds while creating prompts for the test set. The exemplars are sampled from the train set of the test tasks and performance is averaged. We report both the average and the worst-case performances over all five seeds.

\noindent \textbf{Significance Analysis.} For all reported results, we pay special care in determining the significance of the result. If both the average and the worst-case performance are better, we adjudge the result significant (bold numbers) based on lower standard deviation.

\subsubsection{\textbf{Performance on tasks from unseen domains}}
We report the performance of MAML-2-1 and MAML-2-4 along with the baselines as discussed before in Table~\ref{tab:performance-unseen-tasks} (a) and (b) on only unseen domain of tasks. Note that unseen tasks share no task type with the training set as well as sampled from completely disjointed domains. 

\noindent \textbf{Complete Data.} We observe that \textit{MAML-en-LLM} settings outperform MetaICL on \textbf{5 out of the 7 }task settings (win rate of \textbf{0.71}) for standard models and \textbf{4 out of 7} task settings (win rate of \textbf{0.57}) for channel models on \textit{complete data setting}. For the task settings where MAML settings do not outperform MetaICL, we observe that Non-Para$\rightarrow$Para on standard models underperforms significantly (discussed in Subsection~\ref{sec:type-problem}). 

\noindent \textbf{Limited Data.} Similarly, \textit{MAML-en-LLM} outperforms MetaICL on \textbf{4 out of 7} settings (win rate of \textbf{0.57}) using both standard and channel models on \textit{limited data setting}. These results demonstrate the efficacy of our approach in low-resource settings. For most other settings \textit{MAML-en-LLM} performs at par with MetaICL.

\noindent \textbf{Takeaway.} The outperformance of \textit{MAML-en-LLM} on the unseen task setting implies that our method learns a more generalized parameter set. This is a direct consequence of \textit{MAML-en-LLM} exploring a wider set of parameter space as demonstrated in Figure~\ref{fig:maml-motivation}. 

\subsubsection{\textbf{Performance on all tasks}}
Similar to unseen tasks, we report the performances of MAML-2-1 and MAML-2-4 settings in Table~\ref{tab:performance-all-tasks} (a) and (b) along with the associated baselines. 

\noindent \textbf{Complete Data.} \textit{MAML-en-LLM} settings outperform MetaICL on the \textit{complete data settings} for all the tasks on \textbf{4 out of 7} (win rate of \textbf{0.57}) data settings. Channel models perform much better, beating MetaICL on \textbf{5 out of 7} (win rate of \textbf{0.71}) data settings. 

\noindent \textbf{Limited Data.} For the limited data settings, \textit{MAML-en-LLM} settings outperform MetaICL on \textbf{6 out of 7} (win rate of \textbf{0.85}) task settings on both Standard and Channel models. As before, \textit{MAML-en-LLM} settings significantly outperform or are comparable with the MetaICL counterparts on most settings. 

\noindent \textbf{Takeaway.} The results on both \textit{complete} and \textit{limited} task settings show that even though \textit{MAML-en-LLM} performs meta-updates once every $2kn$ batches, it is insensitive to amount of train data. This further attests to \textit{MAML-en-LLM}'s efficacy and wide use cases.

\begin{table*}[t]
\begin{adjustbox}{max width=0.9\textwidth}
\begin{tabular}{cccccccc}
\hline
 Method        & \multicolumn{1}{p{1.5cm}}{\centering HR \\ $\rightarrow$ LR} & \multicolumn{1}{p{1.5cm}}{\centering Class \\ $\rightarrow$ Class} & \multicolumn{1}{p{2cm}}{\centering Non-Class \\ $\rightarrow$ Class} & \multicolumn{1}{p{1.5cm}}{\centering QA \\ $\rightarrow$ QA} & \multicolumn{1}{p{1.5cm}}{\centering Non-QA \\ $\rightarrow$ QA} & \multicolumn{1}{p{1.5cm}}{\centering Non-NLI \\ $\rightarrow$ NLI} & \multicolumn{1}{p{1.5cm}}{\centering Non-Para \\ $\rightarrow$ Para}  \\
\hline
MetaICL  & 42.79 & 38.15 & \textbf{40.49} & 60 & 38.43 & 80.23 & \textbf{61.75}\\
MAML-2-1 & 34.44 & \textbf{42.86} & 34.14 & \textbf{61.25} & 38.75 & \textbf{83.75} & 32.79 \\
MAML-2-4 & \textbf{44.65} & 41.1 & 33.41 & 58.75 & \textbf{40.31} & 66.24 & 34.05 \\
\hline
Channel MetaICL  & 50.82 & \textbf{49.76} & 44.71 & \textbf{57.81} & 46.87 & 44.86 & 55.43 \\
Channel MAML-2-1 & 53.61 & 47.62 & 47.88 & 55.62 & 48.12 & \textbf{60.68} & \textbf{59.86}\\
Channel MAML-2-4 & \textbf{54} & 48.32 & \textbf{48.34} & 56.25 & \textbf{48.12} & 59.28 & 58.44\\
\hline
\end{tabular}
\end{adjustbox}
\caption{Results on very few-shot adaptation on unseen domains of tasks on standard models trained with MetaICL (Row-1), MAML (Rows-2,3) and channel models trained with MetaICL (Row-4), MAML (Rows-5,6).  }
\label{tab:performance-few-shot-adapt}
\end{table*} 

\subsection{Effect of Task Complexity on Exploration States (Number of Tasks)}
\subsubsection{\textbf{Task complexity}} We preface the discussion around exploration states by discussing the exact task settings. Note that we have a total of seven settings out of which - two are classification, one is natural language inference (NLI) which is similar to classification, two are question answering (QA), and one is paraphrasing while the last one is a mixture of all of them characterized only by the amount of data. We designate the set of Classification and NLI tasks as \textit{less-complex} while Paraphrasing and QA are \textit{more-complex}. As a direct consequence to task splits, \textit{more-complex} tasks should ideally require a wider parameter space exploration while \textit{less-complex} tasks should require relatively less exploration. 

\subsubsection{\textbf{Complexity directly affects performance of various exploration states}}
We utilize \textit{MAML-en-LLM} on 2 settings - with 1 task and 4 tasks represented by MAML-2-1 and MAML-2-4, respectively. The more the number of tasks, the more parameter space is explored by the model before performing the meta-update. However, on the flip side, more exploration leads to slower convergence to the minima. The trend we observe is that more number of tasks (MAML-2-4) benefit more-complex settings like QA and Paraphrasing, while a smaller number of tasks (MAML-2-1) help less-complex settings like classification and NLI. (Table~\ref{tab:performance-all-data},\ref{tab:performance-all-small-data}). We report results on the validation set in the Appendix.

\subsubsection{\textbf{Further discussion on performance}}
\label{sec:type-problem}
Even though the task settings are chosen keeping in mind the non-overlap of tasks, not all tasks are created equal. We believe the choice behind the task selection and splits is not explored in detail in \cite{min2022metaicl}. We attempt to shed light on why MetaICL or \textit{MAML-en-LLM} do not perform well on specific tasks like Non-Para$\rightarrow$Para. From our experiments, we observe both MetaICL and \textit{MAML-en-LLM} on standard models actually degrade performance from out-of-box pre-trained models (Refer last column in Table~\ref{tab:performance-all-data}). This observation alludes to the fact that standard model training of MetaICL and \textit{MAML-en-LLM} actually causes some forgetting to the out-of-box pre-trained models by disrupting model weights out of box. In other words, pre-trained out-of-box models are already good enough to perform complex tasks like paraphrasing. Hence, task type and design need to be closely monitored to utilize meta-training approaches.

\subsubsection{\textbf{Further discussion on training sets}}
Non-QA -> QA and Non-Para->Para require meta-training on a relatively easier subset (Non-QA and Non-Para sets mostly include easier classification and NLI tasks), which makes the test sets significantly challenging. Indeed, Meta-training on these tasks using MetaICL even causes some forgetting as compared to pre-trained models. Please refer to Table~\ref{tab:performance-all-data} where RawLM outperforms MetaICL baseline on challenging data settings. Refer Table \ref{tab:performance-all-data} where both Non-Para -> Para (37.26 to 33.1) and Non-QA to QA (39 to 36.71) give a degradation in performance. This behavior is mirrored in the original paper as well \cite{min2022metaicl}.  Hence, the behavior observed using MAML-en-LLM on these particular settings is not any different from MetaICL itself and requires further analysis - out of the scope of this paper.


\subsection{Effect of Optimizer Choice}
Multiple works have demonstrated that optimizers play a crucial role in the effective training of LLMs. Stateless optimizers like Stochastic Gradient Descent (SGD) underperform newer adaptive optimizers like AdamW \cite{loshchilov2017decoupled} in LLM training. Note that MetaICL\cite{min2022metaicl} utilizes AdamW. However, MAML-en-LLM utilizes a dual optimization problem - which requires two separate optimizers for the inner and outer optimizations. In Table~\ref{tab:optim-ablation}, we provide ablation studies with two different optimizers - one stateless (SGD) and one adaptive (AdamW). Due to compute limitations, we sample a 10\% subset of the training and test data across two seeds for both MetaICL and MAML-2-1. In row 4, we report the performance utilizing a combination of optimizers for inner and outer optimizations. Under MAML-2-1, when using stateless optimizers (SGD+SGD), we do not see an increase in performance. Next, as the inner optimizer is re-initialized after every meta-update, we use SGD for inner and AdamW for outer - resulting in a significant increase in performance. Note that using a stateless optimizer in the outer loop is identical to utilizing both stateless optimizers and the performance is identical to SGD+SGD. Lastly, we report results on using adaptive optimizers in both inner and outer optimization steps with and without moment parameter sharing (discussed in subsection \ref{sec:method-optim}).

\begin{table}[h]
\begin{tabular}{cccc}
\hline
\textbf{Method} & \textbf{Optimizer}  & Seed = 10 & Seed = 20  \\
\hline
NoICL & -  & 24.9 & 28.2 \\
RawLM & -  & 35.2 & 28.4 \\
\hline
\multirow{2}{*}{MetaICL} & SGD & 34.6 & 36.2 \\
                         & AdamW & 34.8 & 36.0 \\
\hline
\multirow{5}{*}{MAML-2-1} & SGD + SGD & 36.4 & 35.6 \\
                          & SGD + AdamW & 40.0 & 38.6 \\
                          & AdamW + SGD & 36.4* & 35.6* \\
                          & AdamW + AdamW (d) & 40.0 & 38.6 \\
                          & AdamW + AdamW & \textbf{42.0} & \textbf{42.8} \\
\hline                    
\end{tabular}
\caption{Performance over two seeds for NoICL, RawLM, MetaICL and MAML-2-1 methods utilizing various combinations of optimizers. The `optimizer' column in row 4 uses the notation X+Y, where X and Y are the inner and outer optimizers respectively. Note that (*) represents an identical setting to SGD+SGD. The AdamW+AdamW (d) setting utilizes AdamW without moment sharing. Utilizing adaptive optimizers (AdamW) with parameter sharing in inner and outer optimization yields the best results. (Sampled from Non-NLI $\rightarrow$ NLI)}
\label{tab:optim-ablation}
\end{table}

\subsection{Experiment-2: Very Few Shot Adaptation}
\label{sec:very-few-shot-exp}
The second experiment involves adapting the meta-trained models to unseen domains using only a few samples. \noindent \textbf{Setup:} We utilize the training set of test tasks for sampling both the adaptation samples and their prompts, ensuring that the same target prompt is never utilized in exemplars. We sample 16 exemplars or 256 sequence lengths whichever is lower for the prompts. We consider a total of 16 adaptation data points. The model is adapted using the adaptation training samples for 16 steps (1 pass over all points) with a learning rate of $1$x$10^{-7}$ using AdamW. The testing set remains identical to the unseen test tasks mentioned in Table~\ref{tab:dataset-split}.

\noindent \textbf{Takeaway:} Generalization to unseen tasks is a paramount problem in LLM literature. We report the results of very few-shot adapted models in Table~\ref{tab:performance-few-shot-adapt}.  Note that the starting unadapted standard and channel models are identical to the \textit{complete data setting}. We observe for standard models, \textit{MAML-en-LLM} settings outperform MetaICL on \textbf{5 out of 7} tasks (win rate of \textbf{0.71}). Similarly, channel models outperform MetaICL on \textbf{5 out of 7 }tasks (win rate of \textbf{0.71}). Once again we observe that generalization to unseen domains of tasks is better captured by \textit{MAML-en-LLM} as the model explores wider parameter space during training, thus learning better parameter initializations for adaptation. This behavior has been well documented in Meta-learning literature like \cite{finn2017model,antoniou2018train}.

\subsection{Runtime Analysis}
In this section, we detail some practical considerations to keep in mind during meta-training models using \textit{MAML-en-LLM}. For reference, Figure~\ref{fig:maml-motivation} details the training procedure schematically.
\begin{itemize}
    \item \textbf{Model sizes}: Let us assume the size of the computation graph is in order of the number of parameters of the model in question. Assume $S(\theta) = c*O(\theta) = O(\theta)$ to be the size of the computational graph during training. For MetaICL, as there is no adaptation phase, the gradients are computed only once. Hence at any given instance, the maximum memory utilization of the computational graph is $2*S(\theta)$, where the only values are the present parameter state and the gradients. However, for \textit{MAML-en-LLM}, after the adaptation step the number of parameters in the computational graph are $S(\theta)$ (unadapted params), $k*S(\theta)$ ($k$ is number of tasks), $S(\theta)$ (meta-update) which is a total of $(k+2)*S(\theta)$ parameters. Hence the memory utilization is a linear function of the number of tasks. Thus, it is important to control the number of tasks as per the meta-training strategy and the type of tasks themselves. For our purpose, GPT-2 Medium with 355 million parameters is used.
    \item \textbf{Optimizer states} \label{optimizer_appendix}: \textit{MAML-en-LLM} as opposed to MetaICL utilizes 2 optimizers for the adaptation and the meta-update steps respectively. Even though it might be tempting to utilize state-less optimizers like SGD and Adam, LLM training has been shown to work better when the updates are conditioned on the last gradient state such as with adaptive optimizers like AdamW \cite{loshchilov2017decoupled}. Hence, we utilize AdamW for both adaptation and meta-updates (Optim 1 and Optim 2 in Figure~\ref{fig:maml-motivation}). However, after each update step, we copy the last optimizer state back and forth to both optimizers to smoothen out the training procedure.
\end{itemize}

\section{Limitations}
We acknowledge that our method has a few major limitations in practice. We discuss the limitations in detail below:
\begin{itemize}
    \item \textbf{Unstable Training/Performance:} Due to the dual optimization procedure of MAML-based approaches, the training procedure is bumpy and non-smooth, making the training and validation losses spiky. This observation has been noted by \cite{antoniou2018train}. This makes the training of large models even more unstable and care must be taken while training employing smoothing methods like early stopping, gradient clipping, etc.
    \item \textbf{Sensitive Hyperparameters:} The training procedure is extremely sensitive to learning rate, warmup and decay.
    \item  \textbf{Catastrophic Forgetting:} In some exceptional cases, our method can cause catastrophic forgetting of the model parameters and underfit on complex use-cases, as observed in \cite{min2022metaicl} as well. Further investigation into this behavior is required.
    \item \textbf{Runtime Complexity:} As always, the runtime complexity of MAML is high. We recommend users to carefully consider tradeoffs before utilizing MAML-en-LLM. 
    \item \textbf{Task complexity:} Based on our observations, we recommend utilizing MAML-2-1 for tasks that benefit from faster convergence and low-param space exploration (Classification, NLI) which are less complex than tasks that require larger param space exploration (Paraphrasing, QA)
\end{itemize}

\section{Conclusion}
In this paper, we proposed a novel method \textit{MAML-en-LLM} that meta-trains pre-trained out-of-box models using the principles proposed in meta-learning literature. We demonstrated that \textit{MAML-en-LLM} explores a much wider parameter space than current SOTA meta-training methods like MetaICL and MetaICT due to adaptation to multiple sets of parameters before the actual meta-update. Empirically, \textit{MAML-en-LLM} outperforms MetaICL on both standard and channel models on an extensive set of tasks in both seen and unseen domains. Subsequently, we also demonstrate that models trained using \textit{MAML-en-LLM} can be quickly adapted in a few-shot manner to a set of tasks in the unseen domain. Overall, \textit{MAML-en-LLM} has been empirically demonstrated to outperform MetaICL on performance and generalization. We hope our study motivates the community to utilize classical meta-learning principles in the meta-training of LLMs in the future.

\bibliographystyle{ACM-Reference-Format}
\bibliography{ref}

\newpage
\clearpage
\appendix

\section{Appendix}
\label{sec:appendix}
Table~\ref{tab:val-perf} details the validation performance (computed over 5 seeds) for the \textit{MAML-en-LLM} standard and channel models. The validation test is selected from the training set of the test tasks. We see that the performance agrees with the test tasks.




\begin{table}[h]
\centering
\begin{adjustbox}{max width=0.5\textwidth}
\begin{tabular}{c|c|cccc}
\hline
                    & Seed & MAML-2-1 & MAML-2-4 & Channel-2-1 & Channel-2-4 \\
                    \hline
\multirow{5}{*}{HR $\rightarrow$ LR} &  100    &      72.4 & 70.82 & 70.62 & 71.91     \\
                    &   13   &          69.01 & 64.86 & 69.15 & 69.96           \\
                    &   21   &         67.52 & 63.46 & 73.55 & 67.74               \\
                    &   42   &         68.07 & 65.36 & 72.83 & 70.38        \\
                    &   87  &    70.1 & 62.83 & 73.33 & 73.02                \\
                    \hline
\multirow{5}{*}{Class $\rightarrow$ Class}   &  100    &     48.04& 51.84& 50.48 &61.25    \\
                    &   13   &      47.85 & 47.47& 43.32 &60.95   \\
                    &   21   &     48.36 &48.33& 41.46& 62.17   \\
                    &   42   &    46.5& 46.61 &50.29 &58.59 \\
                    &   87  &      42.64& 40.95& 51.65 &62.71    \\
                    \hline
\multirow{5}{*}{Non-Class $\rightarrow$ Class}   &  100    &    62.39 & 61.2 & 59.99 & 61.91 \\
                    &   13   &     56 &55.38& 59.66& 61.95                \\
                    &   21   &   56.51 &52.08& 61.72& 65.06                \\
                    &   42   &    59.09& 53.84& 58& 63.53               \\
                    &   87  &     57.28 &51.84& 58.81& 60.37            \\
                    \hline
\multirow{5}{*}{QA $\rightarrow$ QA}   &  100    &    78.125 &77.27& 66.19 &83.8    \\
                    &   13   &    78.125& 78.12& 73.01& 80.11                 \\
                    &   21   &     74.71 &66.76 &73.86& 81.25                \\
                    &   42   &      77.55& 71.02 &69.88& 81.81              \\
                    &   87  &     74.43& 69.03& 69.88& 79.26    \\
                    \hline
\multirow{5}{*}{Non-QA $\rightarrow$ QA}   &  100    &    75.56 &76.13& 67.89 &82.67     \\
                    &   13   &      78.4 &76.7& 72.44& 79.54  \\
                    &   21   &    79.54& 72.72& 75.56& 80.39     \\
                    &   42   &     73.86& 67.89 &72.44& 82.95   \\
                    &   87  &    75.85& 65.62& 72.44& 76.7  \\
                    \hline
\multirow{5}{*}{Non-NLI $\rightarrow$ NLI}    &  100    &   32.61 &34.58 &31.16& 46.04    \\
                    &   13   &     34.15& 44.48& 37.43 &44.72   \\
                    &   21   &    45.15 &39.54& 32.53& 49.86\\
                    &   42   &     40.19& 34.07& 26.84& 43.76   \\
                    &   87  &     39.58& 46.74& 43.35& 44.82  \\
                    \hline
\multirow{5}{*}{Non-Para $\rightarrow$ Para}   &  100    &     55.3 &50.55& 38.12 &46.12  \\
                    &   13   &     42.96& 38.03& 45.94 &50.54     \\
                    &   21   &      53.12& 45.52& 32.07& 58.39    \\
                    &   42   &     38.5& 50.39& 43.5 &43.33  \\
                    &   87  &     41.41& 46 &38.94& 64.72   \\
                    \hline
\end{tabular}
\end{adjustbox}
\caption{Validation performance on the training set of test task on Standard and Channel models.}
\label{tab:val-perf}
\end{table}

\end{document}